\documentclass[letter, 10 pt, conference]{ieeeconf}  

\IEEEoverridecommandlockouts 
\usepackage[english]{babel}
\usepackage[utf8]{inputenc}
\usepackage{times} 
\usepackage{cite}

\usepackage{hyperref}
\usepackage[nolist]{acronym}

\usepackage{multirow}

\usepackage{siunitx}
\usepackage{graphicx} 
\usepackage{epsfig} 
\usepackage{pgfplots}
\usepackage{caption}
\usepackage[font=footnotesize]{subcaption}
\usepackage{pgfplots}

\usepackage{float}

\usepackage[export]{adjustbox}

\usepackage{mathtools}
\usepackage{nicefrac}
\usepackage{amsmath}
\usepackage{amssymb}
\usepackage{bm} 

\DeclareMathOperator*{\minimize}{minimize}

\def\xx{\mathsf{x}}
\def\yy{\mathsf{y}}
\def\zz{\mathsf{z}}

\usepackage{etoolbox}
\makeatletter%
\AfterPreamble{%
	\usepackage{hyperref}%
	\let\oldhypertarget\hypertarget%
	\renewcommand{\hypertarget}[2]{%
		\oldhypertarget{#1}{#2}%
		\protected@write\@mainaux{}{%
			\string\expandafter\string\gdef%
			\string\csname\string\detokenize{#1}\string\endcsname{#2}%
		}%
	}%
	\newcommand{\myhyperlink}[1]{%
		\hyperlink{#1}{\csname #1\endcsname}%
	}%
}
\makeatother%

\newcounter{Remark}
\newcounter{Problem}
\usepackage{algorithm}
\usepackage[noend]{algpseudocode}

\makeatletter
\def\BState{\State\hskip-\ALG@thistlm}
\makeatother

\usepackage{tikz}
\pgfdeclarelayer{foreground}
\pgfsetlayers{background, main, foreground}
\usetikzlibrary{quotes, angles, backgrounds, arrows, automata, shapes, positioning, calc, through, spy, decorations.markings}

\usepackage{aircraftshapes} 

\tikzset{
	imglabel/.style={
		rectangle,
		inner sep=2pt,
		text=black,
		minimum height=1em,
		text centered,
		fill=white,
		fill opacity=1.0,
		text opacity=1,
		anchor=south west,
	},
}

\tikzset{
	state/.style={
		rectangle,
		draw=black, very thick,
		minimum height=1.0em,
		text centered,
	},
}

\graphicspath{{./figures/}}

\newcommand\copyrighttext{%
    \small \begin{center} \color{red} \textcopyright\,2023 IEEE. Personal use of this material is permitted. Permission from IEEE must be obtained for all other uses, in any current or future media, including reprinting/republishing this material for advertising or promotional purposes, creating new collective works, for resale or redistribution to servers or lists, or reuse of any copyrighted component of this work in other works. \end{center}}
\newcommand\copyrightnotice{%
	\begin{tikzpicture}[remember picture,overlay]
	\node[anchor=south,yshift=25.6cm] at (current page.south) 
	{\color{red}\fbox{\parbox{\dimexpr\textwidth-\fboxsep-\fboxrule\relax}{\copyrighttext}}};
	\end{tikzpicture}%
}

\title{\copyrightnotice \Large \bf Communications-Aware Robotics: Challenges and Opportunities
}

\author{Daniel Bonilla Licea${^1}$, Giuseppe Silano$^1$, Mounir Ghogho${^2}$, and Martin Saska${^1}$
	\thanks{$^1$Daniel Bonilla Licea, Giuseppe Silano, and Martin Saska are with the Faculty of Electrical Engineering, Czech Technical University in Prague, Czech Republic (emails: {\tt\small \{bonildan, giuseppe.silano, martin.saska\}@fel.cvut.cz}).}
	\thanks{$^2$Mounir Ghogho is with the International University of Rabat, Morocco (email: {\tt\small mounir.ghogho@uir.ac.ma}).}
	\thanks{This work was partially funded by the European Union's Horizon 2020 research and innovation programme AERIAL-CORE under grant agreement no. 871479, by the CTU grant no. SGS23/177/OHK3/3T/13, by the Czech Science Foundation (GAČR) within research project no. 23-07517S, and by the OP VVV funded project CZ.02.1.01/0.0/0.0/16 019/0000765 ``Research Center for Informatics".}
}    

\begin{document}
	
\maketitle
\thispagestyle{empty}
\pagestyle{empty}



\begin{acronym}
	\acro{AoA}[AoA]{Angle of Arrival}
	\acro{AoD}[AoD]{Angle of Departure}
	\acro{AWGN}[AWGN]{Additive White Gaussian Noise}
	\acro{BS}[BS]{Base Station}
	\acro{CaR}[CaR]{Communications-assisted Robotics}
	\acro{CaTP}[CaTP]{Communications-aware Trajectory Planning}
	\acro{FSO}[FSO]{Free Space Optical}
	\acro{FDD}[FDD]{Frequency-Division Duplexing}
	\acro{GTMR}[GTMR]{Generically Tilted Multi-Rotor Model}
	\acro{LoS}[LoS]{Line of Sight}
	\acro{MR}[MR]{Mobile Robot}
	\acro{NMPC}[NMPC]{Nonlinear Model Predictive Control}
	\acro{PD}[PD]{Photo Diode}
	\acro{PV}[PV]{Photovoltaic}
	\acro{RaC}[RaC]{Robotics-assisted Communications}
	\acro{RF}[RF]{Radio Frequency}
	\acro{SIL}[SIL]{Software-in-the-loop}
	\acro{SNR}[SNR]{Signal-to-Noise Ratio}
	\acro{TP}[TP]{Trajectory Planning}
	\acro{UAV}[UAV]{Unmanned Aerial Vehicle}
    \acro{UGV}[UGV]{Unmanned Ground Vehicle}
    \acro{wrt}[w.r.t.]{with respect to}
\end{acronym}



\begin{abstract}

The use of~\acfp{UGV} and~\acfp{UAV} has seen significant growth in the research community, industry, and society. Many of these agents are equipped with communication systems that are essential for completing certain tasks successfully. This has led to the emergence of a new interdisciplinary field at the intersection of robotics and communications, which has been further driven by the integration of~\acp{UAV} into 5G and 6G communication networks. However, one of the main challenges in this research area is how many researchers tend to oversimplify either the robotics or the communications aspects, hindering the full potential of this new interdisciplinary field. In this paper, we present some of the necessary modeling tools for addressing these problems from both a robotics and communications perspective, using the~\ac{UAV} communications relay as an example.

\end{abstract}



\begin{keywords}
    Communications; Control Design; Motion Planning; Multi-Rotor~\acp{UAV}; Communications relay.
\end{keywords}

\section{Introduction}
\label{sec:introduction}

The intersection of communications and robotics has garnered increasing attention, as demonstrated by the growing number of publications in both the robotics~\cite{Gasparri2017TRO, Licea2020TRO} and communications~\cite{Zeng2017TWC, Wu2018TWC} communities. This interest is partly due to the development of 5G and 6G technologies that aim to integrate~\acfp{UAV} into cellular communication networks~\cite{Zeng2019ProceedingsIEEE}. These interdisciplinary problems can be broadly classified into two categories based on the application scenario:~\acf{RaC} and~\acf{CaR}.

In~\ac{RaC} applications, one or more robots are integrated into a communication network to improve their performance. They can serve as mobile relays~\cite{Zeng2019ProceedingsIEEE} or as~\acp{BS} that extend the communication range of the network or provide communication to more network users, respectively. In traditional mobile communication, the position of the transceiver is considered ``uncontrollable" and random. In~\ac{RaC}, the position of the transceiver is ``controllable", representing an additional communication system parameter to be optimized. This difference presents exciting opportunities for the design of communication systems involving robots.

In~\ac{CaR} applications, robots can exchange information with other agents in the field, such as other robots or~\acp{BS}, to facilitate various tasks. Examples of these tasks include exchanging localization and state estimation data in robot swarms~\cite{Ahmad2022swarms}, using information exchange to help robots escape mazes faster~\cite{Jung2010CM}, and transmitting real-time data from surveying robots to a control center~\cite{LindheICRA2010}. To perform these tasks effectively, the robots must consider how their own motion affects the quality of the communication channel.

To achieve success in~\ac{RaC} and~\ac{CaR} applications, it is necessary to have a comprehensive understanding of both communication and robotics. Ignoring either aspect can result in inefficient or infeasible solutions. When only considering communication, a solution may be inefficient in terms of energy consumption or not feasible given the robot's dynamic and physical constraints. On the other hand, focusing only on robotics and ignoring the impact of communication may result in the failure of the robot to complete its task due to unexpected communication failures caused by poor connectivity or lower than expected bit rates. Therefore, it is important to approach these problems with an interdisciplinary perspective to develop functional and effective solutions and fully capitalize on the opportunities presented in this growing research field. This importance has also been recently highlighted in~\cite{CalvoFullana2021IEEECM, CalvoFullana2021RAL}. However, existing literature often lacks an interdisciplinary approach to addressing~\ac{CaR} and~\ac{RaC} applications. While some recent tutorials~\cite{Zeng2019ProceedingsIEEE, Guo2018TRO} have been published on these problems, they tend to oversimplify either the communications or robotics aspects. 


The purpose of this paper is to address the gap in research on the intersection of communications and robotics by highlighting the need for an interdisciplinary approach. To illustrate this point, in Section~\ref{sec:communicationAwareTrajectory}, we examine a general mathematical structure for the problem of~\acf{TP} in the context of~\ac{CaR} and~\ac{RaC} applications, referred to as~\acf{CaTP}. Then, in Section \ref{sec:uavCommunicationRelay}, we provide an example of a~\acf{CaTP} problem by discussing our work on~\acp{UAV} acting as \textit{communication relays} between two agents. Finally, we outline several research opportunities and challenges of this exciting research area in Section~\ref{sec:researchOpportunities}.



\section{Communications-aware Trajectory Planning}
\label{sec:communicationAwareTrajectory}

The goal of~\ac{CaTP} optimization problems is to find trajectories that take into account both the communications (e.g., communication energy consumption, quality of the communications link, number of transmitted bits) and the robotics aspects (e.g., kinematic and dynamic constraints, motion energy consumption, physical actuation limits). The general structure of a~\ac{CaTP} optimization problem that seeks to determine the predetermined trajectory of a single robot can be expressed as:
\begin{subequations}\label{TPA:eq:1}
    \begin{align}
    &\minimize_{\mathbf{u}, \, \mathbf{x}, \, \mathcal{C}, \, \mathcal{T} } \; \;  { J(\mathbf{u}, \mathbf{x}, \mathcal{C},\mathcal{T}) } \label{subeq:optimizationTarget}\\
    &\quad\;\; \text{s.t.}~\quad \textrm{motion model}, \label{subeq:motionModel} \\
    &\quad\;\;\;\;\qquad \text{channel model}, \label{subeq:channelModel}\\
    &\quad\;\;\;\;\qquad \text{trajectory constraints}, \label{subeq:trajectoryConstraints} \\
    &\quad\;\;\;\;\qquad \text{communications constraints}, \label{subeq:communicationConstraints}
    \end{align}
\end{subequations}
where $\mathbf{u}$ and $\mathbf{x}$ are the robot's control signal and system state, respectively; $\mathcal{C}$ is the set of all communications-related parameters to be optimized, such as modulation order or transmission power\footnote{Note that if all of the communications parameters are fixed, then $\mathcal{C}=\emptyset$.}; and $\mathcal{T}$ is the set of the remaining parameters to be optimized that are not directly related to the communication system, such as the completion duration of the trajectory.

The~\ac{CaTP} optimization problem~\eqref{TPA:eq:1} consists of five elements: the optimization target~\eqref{subeq:optimizationTarget}, which is given by $J(\mathbf{u}, \mathbf{x}, \mathcal{C}, \mathcal{T})$; the motion model~\eqref{subeq:motionModel}, which describes how the robot's state $\mathbf{x}$ evolves based on the control signal  $\mathbf{u}$; the wireless channel model~\eqref{subeq:channelModel}, which describes how the received and transmitted signals behave depending on the robot's position and orientation; generic trajectory constraints~\eqref{subeq:trajectoryConstraints}, such as maximum velocity and acceleration for the robot; and finally, constraints~\eqref{subeq:communicationConstraints} related to the communication system performance and/or goals. We now discuss these five elements in more detail. 

\textbf{Optimization target}: the optimization target of~\ac{CaTP} problems can vary, but it can generally be divided into three categories: those that include both communications and robotics terms, those that only consider robotics aspects, and those that only consider communication aspects. A common robotics-related term used in the cost function~\eqref{subeq:optimizationTarget} of~\ac{CaTP} problems is the energy consumed by the robot while tracking the desired trajectory, i.e., $\lVert \mathbf{u}^\top \mathbf{u} \rVert^2$. Another common robotics-related term is the distance traveled by the robot. This assumes that the energy consumption due to motion can be accurately obtained from the distance traveled by the robot. Another common robotics-related metric is the total time that the robot takes to complete the designed trajectory, i.e., $\Delta t = t_f - t_i$, where $t_f$ and $t_i$ represent the end and start times of the robot's motion, respectively. As for the communication-related term $\mathcal{C}$ in the optimization target~\eqref{subeq:optimizationTarget}, typical examples include the energy spent by the robot's communication system, the total number of bits transmitted, and the amount of time during which the robot was disconnected from the communication network. Alternatively, the optimization target $J(\mathbf{u}, \mathbf{x}, \mathcal{C}, \mathcal{T})$ is constant in certain problem formulations, resulting in an optimization problem~\eqref{TPA:eq:1} that becomes a constraint satisfaction problem. The solution to this problem would then yield a trajectory (or set of trajectories) that satisfies all constraints.

\textbf{Motion model}: the motion model in~\ac{CaTP} optimization problems describes how the robot's state $\mathbf{x}$ changes based on the control signal $\mathbf{u}$. This can be represented as either a continuous or discrete function, such as $\dot{\mathbf{x}} = \mathbf{f}(\mathbf{x}, \mathbf{u})$ or $\mathbf{x}_{k+1} = \mathbf{f}(\mathbf{x}_k, \mathbf{u}_k)$\footnote{ $\mathbf{x}_k$ and $\mathbf{x}_{k+1}$ represent the next and current states of the discrete-time dynamical system, respectively.}. The motion model can include kinematic or dynamic models of the robot, which may be nonlinear with respect to the control signal $\mathbf{u}$. While nonlinear models can provide more accurate results, they may be more complex to work with. Linearizing the motion model can simplify the control and optimization process, though the resulting solution may be suboptimal~\cite{Siciliano2016Handbook}.

\textbf{Channel model}: the channel model specifies how the signals that are transmitted and received by the robot vary based on the robot's position and orientation (also known as its pose) and the physical surroundings in which it is communicating. This includes whether the environment is indoors or outdoors, urban or rural, and so on. The channel model is often based on a statistical model of the wireless channel, which can be either deterministic or stochastic depending on the amount of information available about the channel. It is used to predict the strength of the signal or the~\ac{SNR} at the receiver based on the transmitter and receiver positions, the frequency of the transmission, and other factors, such as shadowing or multipath fading~\cite{HammoutiUnet2018, uav15, UAVChannelsurvey2}. Additionally, the channel model is used to determine the required transmission power or modulation scheme needed to achieve a certain level of performance. The channel model plays a crucial role in the~\ac{CaTP} problem because it determines the feasibility and performance of the communication links between the robot and other nodes, which in turn affects the robot's ability to perform its tasks.

\textbf{Trajectory constraints}: there are three categories of optimization targets in~\ac{CaTP} problems. The first category includes constraints related to both communications and robotics (e.g., maximum robot velocity and acceleration, and maximum bandwidth for the communication channel). The second category includes exclusively robotics-related constraints (e.g., maximum robot velocity and acceleration) with the third category including only communications-related constraints (e.g., maximum bandwidth for the communication channel). The specific constraints set up can vary based on the requirements of the mission.


\textbf{Communications constraints}: these affect the overall performance of the communication system or the wireless channel experienced by the robot along its entire trajectory. When using stochastic channel models, the constraints related to the wireless channel are probabilistic. For example, we can require that the robot transmit an average of at least $N$ bits along its trajectory, as expressed by the following constraint:
\begin{equation}\label{TPA:eq:3}
\int_{0}^T \mathbb{E} [R(\mathrm{SNR}(\mathbf{p}_1(t), \mathbf{p}_2(t), t))]\mathrm{d}t\geq N,
\end{equation}
where $\mathrm{SNR}(\mathbf{p}_1(t), \mathbf{p}_2(t), t)$ is the instantaneous~\ac{SNR} observed at time instant $t$ for the link between the robot at position $\mathbf{p}_1(t)$ and to another node at $\mathbf{p}_2(t)$. The function $R(\cdot)$ determines the bit rate transmission as a function of the instantaneous~\ac{SNR}, depending on the modulation scheme. The expected value in~\eqref{TPA:eq:3} is taken~\ac{wrt} all possible realizations of the wireless channel. It is important to note that due to the stochastic nature of the wireless channel, the constraint~\eqref{TPA:eq:3} does not guarantee that the robot will always be able to transmit at least $N$ bits, but rather that, on average, robots following the same trajectory in the same environment will transmit at least $N$ bits. Additionally,~\eqref{TPA:eq:3} assumes that the transmission process is continuous (as indicated by the integral and the fact that the integrand is treated as a continuous function of time). Although the transmission process may not be continuous (data may be transmitted in packets rather than as a continuous, uninterrupted flow), we can still assume that it is continuous for simplicity, since the dynamics of the motion process are generally much slower than the dynamics of the transmission process.

There are also constraints that directly affect the wireless channel. One advantage of directly constraining the wireless channel is that it makes the optimization problem independent of the specific communication system being used (e.g. modulation type). One common type of constraint has the following form:
\begin{equation}\label{TPA:eq:4}
\mathrm{Pr}\left(\mathrm{SNR}(\mathbf{p}_1(t), \mathbf{p}_2(t), t)\geq \gamma_0\right)\geq 1-\varepsilon,\ \forall\ t\in[0,T],
\end{equation}
where $\mathrm{Pr}\left(\mathrm{SNR}(\mathbf{p}_1(t), \mathbf{p}_2(t), t)\geq \gamma_0\right)$ is the probability that $\mathrm{SNR}(\mathbf{p}_1(t), \mathbf{p}_2(t), t)\geq \gamma_0$ occurs. This constraint ensures that the robot's communication system experiences an~\ac{SNR} greater than $\gamma_0$ with a probability of $1-\varepsilon$ throughout the entire trajectory. If the minimum~\ac{SNR} required for the robot to establish communication is $\gamma_0$, then satisfying the constraint~\eqref{TPA:eq:4} ensures that the robot stays connected throughout the trajectory with a probability of $1-\varepsilon$.

Not all five elements in~\eqref{TPA:eq:1} are necessary for all~\ac{CaTP} optimization problems, however the robot motion model (which may not be explicitly mentioned if it is simple enough), the channel model, and the optimization target are essential. As for the constraints, there are three possible scenarios:

\begin{enumerate}
    
    \item \textit{Case 1}: the optimization target involves both communications and robotics aspects. In this case, we may not need any trajectory or communications constraints;

    \item \textit{Case 2}: the optimization target only involves robotics aspects. For example, the optimization target may consider only the motion energy of the robot (i.e., the communications energy is considered negligible compared to the motion energy). In this case, the~\ac{CaTP} problem includes communications constraints and has a reduced set of trajectory constraints, such as only the initial and final positions of the robot;

    \item \textit{Case 3}: The optimization target only involves communications aspects. In this case, the~\ac{CaTP} problem may not have any communication-related constraints.

\end{enumerate}

The constraints discussed previously are \textit{hard constraints}, meaning they must either be satisfied or not. In some cases, these types of constraints can cause issues in the optimization problem. One such issue is the infeasibility of the optimization problem~\eqref{TPA:eq:1}. There may be certain values of the parameters that make it impossible to satisfy some of the hard constraints of the~\ac{CaTP} problem, rendering it infeasible. This can be particularly challenging to identify in complex~\ac{CaTP} problems with many parameters and complicated constraints. Another issue is related to the performance of the solution for the optimization problem~\eqref{TPA:eq:1}. It may happen that a trajectory slightly violates a hard constraint in~\eqref{TPA:eq:1} and achieves a significantly lower value in the optimization target compared to the actual optimal trajectory that satisfies all constraints. These issues can be addressed by relaxing some hard constraints and turning them into \textit{soft constraints}~\cite{boyd_vandenberghe_2004}.

Let us consider the inequality constraint $g(\mathbf{u}, \mathbf{x}, \mathcal{C}, \mathcal{T}) \leq 0$. There are two ways to relax this constraint. The methods to relax equality constraints are similar and only require slight modifications. The first method involves converting the hard constraint into a penalty term that is added to the optimization target in~\eqref{TPA:eq:1}. One way to represent this penalty term is as follows:
\begin{equation}\label{TPA:eq:5}
\mathcal{P}_f = \mu_1\exp( \mu_2 \, g( \mathbf{u}, \mathbf{x}, \mathcal{C}, \mathcal{T})),
\end{equation}
where $\mu_1,\mu_2>0$ are large constants. The term~\eqref{TPA:eq:5} imposes a heavy penalty on the optimization target when the constraint $g(\mathbf{u}, \mathbf{x}, \mathcal{C}, \mathcal{T})\leq 0$ is not satisfied, i.e., when $g(\mathbf{u}, \mathbf{x}, \mathcal{C}, \mathcal{T})>0$. If the constraint $g(\mathbf{u}, \mathbf{x}, \mathcal{C}, \mathcal{T})\leq 0$ is upheld, then $\mathcal{P}_f\approx 0$. We have used this in~\cite{Licea2020TRO} to avoid obstacles in a~\ac{CaTP} problem.

The second method involves the use of slack variables~\cite{boyd_vandenberghe_2004}. To relax the inequality constraint mentioned above~\eqref{TPA:eq:5}, this method reshapes the constraint to $g(\mathbf{u}, \mathbf{x}, \mathcal{C}, \mathcal{T})\leq s_g$, with $s_g\geq 0$, and adds the term $k_s s_g^2$ to the optimization target in~\eqref{TPA:eq:5}, with $k_s \geq 0$. The slack variable $s_g$  allows for a slight violation of the constraint $g(\mathbf{u}, \mathbf{x}, \mathcal{C}, \mathcal{T})\leq 0$. The degree of this violation is controlled by the parameter $k_s$. It is worth noting that when $k_s=0$, the constraint becomes an hard constraint.

Some hard constraints in~\eqref{TPA:eq:5} are related to desired performance and are usually imposed by the designer. For instance, constraints of this type might involve the time needed to complete the mission, the final position of the robot, or the minimum number of bits to transmit. These types of constraints can be relaxed if the specific application, for which the~\ac{CaTP} problem is being solved, permits the relaxation. On the other hand, there are hard constraints that are imposed by the physical world. Examples include the dynamics of the robot, its kinematic or dynamic model, the presence of obstacles, or the maximum transmission power. Relaxing this type of constraint might result in solutions that will not work in the real world.




\section{UAV Communications Relays}
\label{sec:uavCommunicationRelay}

\begin{figure}[tb]
    \centering
    \includegraphics[scale=1]{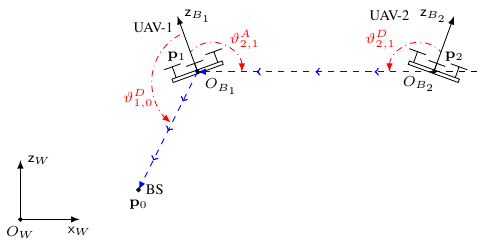}
	\vspace{-0.5em}
	\caption{\ac{UAV}-$1$ serving as a communications relay between the~\ac{UAV}-$2$ and the~\ac{BS}.}
	\label{fig:scenario}
\end{figure}

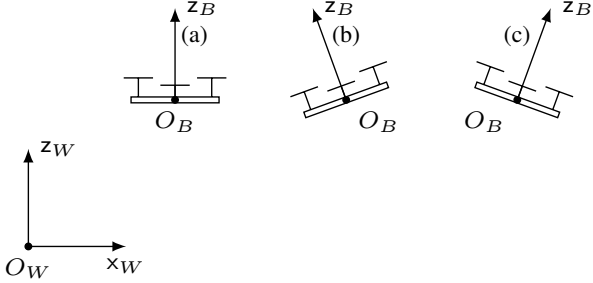
\begin{figure}[tb]
    \centering
    \scalebox{1.3}{
    \begin{tikzpicture}
	    \tikzset{->-/.style={decoration={
            markings,
            mark=at position #1 with {\arrow{>}}}, postaction={decorate}}
        }

	    \node (UAV0) [quadcopter side, fill=white, draw=black, minimum width=1.5cm, rotate=0] at (-2.5,0) {};
        \draw[-latex] (UAV0) node[below]{\scriptsize $O_B$} -- ($ (UAV0) + (0, 0.939692621)$) node[right]{\scriptsize $\zz_B$};
        \node (circle) at (UAV0) [circle, draw, scale=0.2, fill=black] {};
	    \node at (-2.30,0.65) [text centered]{\scriptsize (a)};
	    
	    \node (UAV1) [quadcopter side, fill=white, draw=black, minimum width=1.5cm, rotate=20] at (-0.750,0) {};
        \draw[-latex] (UAV1) node[below right]{\scriptsize $O_B$} -- ($ (UAV1) + (-0.342020143, 0.939692621)$) node[right]{\scriptsize $\zz_B$};
        \node (circle) at (UAV1) [circle, draw, scale=0.2, fill=black] {};
	    \node at (-0.750,0.65) [text centered]{\scriptsize (b)};
	    
		\node (UAV2) [quadcopter side, fill=white, draw=black, minimum width=1.5cm, rotate=-20] at (1.00,0) {}; 
        \draw[-latex] (UAV2) node[below left]{\scriptsize $O_B$} -- ($ (UAV2) + (0.342020143, 0.939692621)$) node[right]{\scriptsize $\zz_B$};
        \node (circle) at (UAV2) [circle, draw, scale=0.2, fill=black] {};
		\node at (1.00,0.65) [text centered]{\scriptsize (c)};
		
		\draw[-latex] (-4,-1.5) node[below]{\scriptsize $O_W$} -- (-4,-0.5) node[right]{\scriptsize $\zz_W$};
		\draw[-latex] (-4,-1.5) -- (-3,-1.5) node[below]{\scriptsize $\xx_W$};
		\node (circle) at (-4,-1.5) [circle, draw, scale=0.2, fill=black] {};
	\end{tikzpicture}
	}
    \vspace{-0.5em}
    \caption{(a) quadrotor hovering, (b) quadrotor moving to the left, and (c) quadrotor moving to the right.}
    \label{fig:UAVmovement}
\end{figure}

To demonstrate the importance of considering both the robotics and communication aspects in~\ac{CaTP} problems, we will consider the scenario of a robotic relay system between a~\ac{BS} and another~\ac{UAV}. In particular, we will focus on the case of a multi-rotor~\ac{UAV}, specifically a quadrotor. As shown in Fig.~\ref{fig:scenario}, both quadrotors are equipped with a single antenna each and need to communicate while moving. One of the key challenges in this problem is the fact that the quadrotors must tilt in order to move, as depicted in Fig.~\ref{fig:UAVmovement}. This motion causes the angular orientation between the antennas to change as the~\acp{UAV} move. As a result, the antenna gain experienced by the~\ac{UAV} can be significantly affected by its motion, depending on the antenna directivity.

\subsection{Motion planning}
\label{sec:motionPlanning}

As stated in Sec.~\ref{sec:uavCommunicationRelay}, the communication system consists of two communications links:~\ac{UAV}-$2$$\rightarrow$\ac{UAV}-$1$ and~\ac{UAV}-$1$$\rightarrow$\ac{BS}. Both~\acp{UAV} are equipped with a single antenna, specifically a half-wave dipole antenna. However, this does not limit the generality of the proposed approach, as it can be easily extended to other types of antennas.

Let us consider a discrete-time dynamic model $\mathcal{H}$ for the quadrotor and denote the~\ac{UAV} sampling period as $T_s \in \mathbb{R}_0$ and the trajectory time as $T \in \mathbb{R}_0$. The time interval is represented by the vector $\mathbf{t}=[0, T_s, \cdots, NT_s]^\top \in \mathbb{R}^{N+1}$, where $\prescript{k}{}{t} = kT_s$, with $k \in \mathbb{N}_0$ and $NT_s=T$, is the $k$-th element of the vector $\mathbf{t}$. Therefore, we can define the state $\mathbf{x}$ and control $\mathbf{u}$ sequences as $\prescript{k}{}{\mathbf{x}} = [\prescript{k}{}{\mathbf{p}}^{(1)}, \prescript{k}{}{\mathbf{v}}^{(1)}, \prescript{k}{}{\mathbf{p}}^{(2)}, \prescript{k}{}{\mathbf{v}}^{(2)}, \prescript{k}{}{\mathbf{p}}^{(3)}, \prescript{k}{}{\mathbf{v}}^{(3)}]^\top$ and $\prescript{k}{}{\mathbf{u}} = [\prescript{k}{}{\mathbf{a}}^{(1)}, \prescript{k}{}{\mathbf{a}}^{(2)}, \prescript{k}{}{\mathbf{a}}^{(3)}]^\top$, where $\prescript{k}{}{\mathbf{p}}^{(j)}$, $\prescript{k}{}{\mathbf{v}}^{(j)}$, and $\prescript{k}{}{\mathbf{a}}^{(j)}$, with $j=\{1, 2, 3\}$, represent the position, velocity, and acceleration of the vehicle at time instant $k$ along the $j$-axis of the inertial frame $O_W$, respectively (see Fig.~\ref{fig:scenario}).

To come up with a trajectory that satisfies the vehicle constraints, while guaranteeing rapid generation and feasibility check, the motion primitives in~\cite{Silano2021RAL} have been considered. Hence, the~\ac{UAV} translational dynamics $[\prescript{k+1}{}{\mathbf{p}}^{(j)}, \prescript{k+1}{}{\mathbf{v}}^{(j)}, \prescript{k+1}{}{\mathbf{a}}^{(j)}]^\top$ can be approximated separately along each $j$-axis with the motion primitives~\cite{Silano2021RAL} as:
\begin{equation}\label{eq:splines}
	\resizebox{0.91\columnwidth}{!}{$
		\begin{bmatrix}
		\frac{\alpha}{120} \prescript{k}{}{t}^5 + \frac{\beta}{24} 
		\prescript{k}{}{t}^4 + \frac{\gamma}{6}
		\prescript{k}{}{t}^3 + \prescript{k}{}{\mathbf{a}}^{(j)} 
		\prescript{k}{}{t}^2 + \prescript{k}{}{\mathbf{v}}^{(j)} 
		\prescript{k}{}{t} + \prescript{k}{}{\mathbf{p}}^{(j)} \\
		\frac{\alpha}{24} \prescript{k}{}{t}^4 + \frac{\beta}{6} 
		\prescript{k}{}{t}^3 + \frac{\gamma}{2} 
		\prescript{k}{}{t}^2 + \prescript{k}{}{\mathbf{a}}^{(j)} 
		\prescript{k}{}{t} + \prescript{k}{}{\mathbf{v}}^{(j)} \\
		\frac{\alpha}{6} \prescript{k}{}{t}^3 + \frac{\beta}{2} 
		\prescript{k}{}{t}^2 + \gamma \prescript{k}{}{t} + 
		\prescript{k}{}{\mathbf{a}}^{(j)}
		\end{bmatrix},
		$}
\end{equation}
where $\{ \alpha, \beta, \gamma \} \in \mathbb{R}$ are design parameters that determine the behavior of the system at the start and end points.

To emphasize the effect of the coupling between the~\acp{UAV}' tilt and its antenna orientation, the following simplifications were performed: (i)~\ac{LoS} is assumed for both communications links, (ii) small-scale fading is neglected, and (iii) it is assumed that the~\ac{BS} uses beamforming to track~\ac{UAV}-$1$. Under these assumptions, the communication channels are modeled using the free space model in~\cite{Licea2020TRO}, which takes into account the effect of the antennas' radiation pattern.

The task objective can be formulated as finding the optimum~\ac{UAV}-$1$ trajectory ($\mathbf{p}_1, \mathbf{v}_1, \mathbf{a}_1$) so as to maximize the number of bits transmitted from~\ac{UAV}-$2$ to the~\ac{BS} via~\ac{UAV}-$1$. Hence, we can write:
\begin{subequations} \label{eq:optimizationProblemNew}
	\begin{align}
		&\hspace{-1.015em}\minimize_{\mathbf{p}_1, \mathbf{v}_1, \mathbf{a}_1}
		\displaystyle\sum_{k=0}^N \Biggl({ \frac{1}{\log_2^p \Bigl(1 + 
		\prescript{k}{}{\xi}_{1,0} \Bigr)}}, { \frac{1}{\log_2^p \Bigl(1 + 
		\prescript{k}{}{\xi}_{2,1} \Bigr)}} \Biggr)^{1/p} \\
		&\hspace{-1.015em}\quad \,\; \text{s.t.}~\quad \prescript{k}{}{\xi}_{1,0} = \frac{D_B^2 \, D^2 
		\mathbf{g}(\vartheta_1, \varphi_1)^\top \mathbf{v}_{1,0} P}{\lVert 
		\prescript{k}{}{\mathbf{p}}_1 - \mathbf{p}_0 \rVert ^2\sigma^2_0},\\
		&\hspace{-1.015em}\qquad\qquad \prescript{k}{}{\xi}_{2,1} = \frac{D^4 \mathbf{g}(\vartheta_1, 
		\varphi_1)^\top \mathbf{v}_{2,1} \mathbf{g}(\vartheta_2, \varphi_2)^\top \mathbf{v}_{2,1} 
		P} {\lVert{ \prescript{k}{}{\mathbf{p}}_2 - \prescript{k}{}{\mathbf{p}}_1 
		\rVert^2 \, \sigma^2_1}},\\
		&\hspace{-1.015em}\qquad \;\;\,~\quad\; \lvert \prescript{k}{}{v}^{(j)} \rvert \leq 
            \bar{v}, \lvert \prescript{k}{}{a}^{(j)} \vert  \leq \bar{a}, \\
		&\hspace{-1.015em}\,\, \qquad \qquad \text{eq.}~\eqref{eq:splines}, \forall k=\{0,1, \dots, N-1\}, \\
		&\hspace{-1.015em}\,\, \qquad \qquad \text{with} \ \vartheta_1, \varphi_1 = 0,
	\end{align}
\end{subequations}
where $P$ is the power of the transmitted signal, $\left(\nicefrac{1}{\log_2^p (1 + 
\prescript{k}{}{\xi}_{1,0})}, \nicefrac{1}{\log_2^p (1 + \prescript{k}{}{\xi}_{2,1})}\right)$ is the smooth approximation of the $\max$ function of the normalized upper bound for the end-to-end data bit rate at discrete time $k$, and $D_B^2 D^2 \mathbf{g}(\vartheta_1, \varphi_1)$ and $D^4 \mathbf{g}(\vartheta_1, \varphi_1)$ are the antenna power gains of the~\ac{UAV}-$1$$\rightarrow$\ac{BS} and~\ac{UAV}-$2$$\rightarrow$\ac{UAV}-$1$ communication links, respectively. The optimization problem includes constraints on the maximum velocity $\bar{v}$ and acceleration $\bar{a}$ of the vehicle to ensure compliance with its motion constraints. Finally, the pitch and roll angles of the $i$-th~\ac{UAV} are represented by $\vartheta_i$ and $\varphi_i$, respectively. The proposed problem formulation~\eqref{eq:optimizationProblemNew} follows the one presented in~\eqref{TPA:eq:1}, including both communication and robot dynamics constraints in the optimization problem.


To validate the proposed approach, numerical simulations in MATLAB and in the Gazebo robotics simulator~\cite{HertICUAS2022} were performed for a power tower inspection task~\cite{Silano2021RAL, SilanoICUAS2021, CalvoICUAS2022-BT}. Videos with the Gazebo simulations can be found at~\url{http://mrs.felk.cvut.cz/optimum-trajectory-relay}, while numerical simulations are available in~\cite{Licea2021EUSIPCO}.



\subsection{Control design}
\label{sec:controlDesign}

This section addresses the robotic relay system problem from a control perspective by designing a~\ac{NMPC} strategy that integrates communication requirements into both the optimization target and constraints. The control strategy is inspired by prior work~\cite{Andriy2022ICUAS, CataffoSMC2022} and is intended to be applicable to a wide range of multi-rotor vehicles, including coplanar designs like quadrotors and fully-actuated platforms with tilted propellers. To this end, we use a~\ac{GTMR} model~\cite{Michieletto2018TRO} equipped with a half-wave dipole antenna to describe the dynamics of the multi-rotor. The control strategy also takes into account realistic actuation limits at the torque level, as well as communication constraints to maximize the number of bits transmitted between~\ac{UAV}-2 and the~\ac{BS}.

The proposed approach incorporates the nonlinear dynamics of the system and realistic physical limitations on the actuators (e.g., rotor acceleration bounds) for agile robot control, but with increased complexity compared to the previous formulation (see Sec.~\ref{sec:motionPlanning}). The use of a~\ac{GTMR} model also allows us to consider the relative orientation of the propellers and the number of rotors $n$ in order to describe the behavior of both under-actuated and fully-actuated platforms. This approach is not platform dependent, meaning it does not rely on specific motion primitive (as in Sec~\ref{sec:motionPlanning}) for the vehicles we use.

\begin{figure}[tb]
    \centering
    \scalebox{1}{
    \begin{tikzpicture}
		\draw[-latex] (-3.25,-1.00) node[below]{$O_W$}  -- (-2.25,-1.00) node[below]{$\xx_W$}; 
		\draw[-latex] (-3.25,-1.00) -- (-3.25,0.00) node[left]{$\zz_W$};; 
		\draw[-latex] (-3.25,-1.00) -- (-2.75,-0.50) node[above]{$\yy_W$};; 
		
		\node (quadrotor) at (0, 0.75) {\includegraphics[scale=1.75]{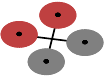}};
        
		\draw[-latex] (-0.175,0.35) arc (55:-15:0.35cm); 
		\draw[-latex] (0.975,1.00) arc (55:-15:0.35cm); 
		\draw[latex-] (0.40,1.75) arc (-25:-120:0.30cm); 
		\draw[latex-] (-0.75,1.15) arc (-25:-120:0.30cm); 
		
		\draw[-latex] (-0.175,0.05) coordinate -- (-0.175,0.55); 
		\draw (-0.075,0.50) node[below right]{$\Omega_1$};
		\draw[-latex] (0.975,0.70) coordinate -- (0.975,1.20);
		\draw (1.15,1.10) node[below right]{$\Omega_2$};
		\draw[-latex] (0.175,1.45) coordinate -- (0.175,1.95);
		\draw (0.175,1.95) node[right]{$\Omega_3$};
		\draw[-latex] (-0.975,0.95) coordinate -- (-0.975,1.45);
		\draw (-0.975,1.25) node[left]{$\Omega_4$};
		
		\draw[-latex] (0,0.75)  -- (1.25,1.25) node[above]{$\yy_B$}; 
		\draw[-latex] (0,0.75) -- (0,2.00) node[left]{$\zz_B$};; 
		\draw[-latex] (0,0.75) -- (1,0) node[below right]{$\xx_B$};; 
		\draw (-0.35,0.85) node[below]{$O_B$}; 
		
    \end{tikzpicture}
    }
    \vspace{-0.75em}
    \caption{A schematic representation of a~\ac{GTMR} model in the case of four propellers ($n=4$) and coplanar orientation.}
    \label{fig:sampleGenericallyTiltedMultiRotor}
\end{figure}
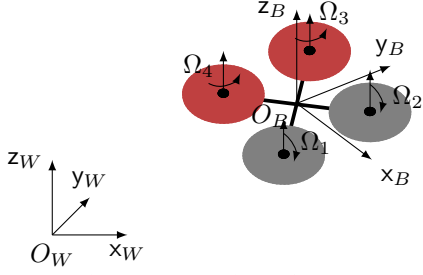

The model~\cite{Michieletto2018TRO} describes a nonlinear dynamic system $\dot{\mathbf{x}} = \mathbf{f} ( \mathbf{x}, \mathbf{u} )$, with state $\mathbf{x} = [ \mathbf{p}^\top \mathbf{q}^\top \mathbf{v}^\top \bm{\omega}^\top ]^\top \in \mathbb{R}^3 \times \mathbb{S}^3 \times \mathbb{R}^6$ and control input $\mathbf{u} \in \mathbb{R}^n$, specifically the propeller spinning velocity $\bm{\Omega} = [\Omega_1, \Omega_2, \cdots, \Omega_n]^\top \in \mathbb{R}^n$. The variables $\mathbf{p} \in \mathbb{R}^3$, $\mathbf{q} \in \mathbb{S}^3$, $\mathbf{v} \in \mathbb{R}^3$, and $\bm{\omega} \in \mathbb{R}^3$ represent the position, orientation, and the linear and angular velocities of the vehicle, respectively. To address the limited bandwidth of the control action, we can extend the model by considering the time derivative of the propeller spinning velocity $\dot{\bm{\Omega}} = [\dot{\Omega}_1, \dot{\Omega}_2, \cdots, \dot{\Omega}_n]^\top \in \mathbb{R}^n$. Hence, the system model~\cite{Michieletto2018TRO} can be rewritten as $\dot{\bar{\mathbf{x}}} = \mathbf{f}(\bar{\mathbf{x}}, \bar{\mathbf{u}})$, where $\bar{\mathbf{x}} \coloneqq [ \mathbf{p}^\top \mathbf{q}^\top \mathbf{v}^\top \bm{\omega}^\top \mathbf{u}^\top ]^\top$ and $\bar{\mathbf{u}} = \dot{\mathbf{u}}$. A schematic representation of the system is depicted in Fig.~\ref{fig:sampleGenericallyTiltedMultiRotor}.

\begin{figure}[tb]
    \centering
    \includegraphics[width=0.8\columnwidth]{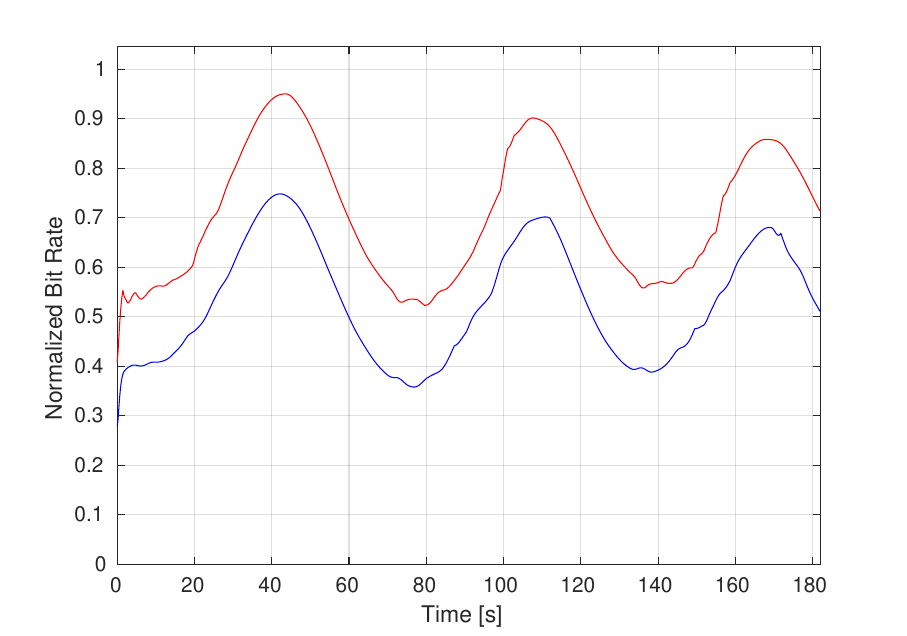}
    \vspace{-0.75em}
    \caption{Normalized bit rate for an illustrative example scenario where the~\ac{UAV} communications relay problem is solved using both a trajectory generation approach (in blue) and a control perspective (in red).}
    \label{fig:comparisonBitTransmitted}
\end{figure}

Thus, the optimal control problem over a prediction horizon of $N$ steps, where $N \in \mathbb{N}_{>0}$ and considering the revised system model equations $\dot{\bar{\mathbf{x}}} = \mathbf{f}( \bar{\mathbf{x}}, \bar{\mathbf{u}} )$, can be formulated as a minimization problem per each time step $t_k = kT_s$\footnote{Due to space constraints, an abuse of notation is made.}, with $T_s$ being the sampling time and $k \in \mathbb{N}_{>0}$, as follows:
\begin{subequations}\label{eq:NMPC_formulation}
    \begin{align}
    &\minimize_{\bar{\mathbf{x}}, \, \bar{\mathbf{u}} } \;\;
    { \sum\limits_{k=0}^N \lVert \mathbf{y}_{\mathrm{d},k} - \mathbf{y}_k \rVert^2_{\mathbf{Q}} } \label{subeq:objectiveFunction} \\
    %
    &\quad \text{s.t.}~\; \bar{\mathbf{x}}_0 = \bar{\mathbf{x}}(\mathbf{t}_k), k = 0,  \label{subeq:stateEquation} \\
    &\;\;\; \quad \quad \bar{\mathbf{x}}_{k+1} = \mathbf{f}(\bar{\mathbf{x}}_k, \bar{\mathbf{u}}_k), k \in \{0, N-1\} , \label{subeq:sysDynamic} \\
    &\;\;\; \quad \quad \mathbf{y}_k = \mathbf{h}(\bar{\mathbf{x}}_k, \bar{\mathbf{u}}_k), k \in \{0, N\}, \label{subeq:outputMap} \\
    &\;\;\; \quad \quad\underline{\gamma} \leq \mathbf{u}_k \leq \bar{\gamma}, k \in \{0, N\}, \label{subeq:uBound} \\
    &\;\;\; \quad \quad \underline{\dot{\gamma}} \leq \bar{\mathbf{u}}_k \leq \bar{\dot{\gamma}}, k \in \{0, N-1\} \label{subeq:dotuBound}, \\
    &\;\;\; \quad \quad \mathbf{g}(\mathbf{u}_k, \mathbf{x}_k, \mathbf{y}_{\mathrm{d},k}, \mathcal{T}) > 0 \label{sueq:misAligConstr}, 
    \end{align}
\end{subequations}
where~\eqref{subeq:objectiveFunction} is the objective function,~\eqref{subeq:stateEquation} sets the initial state conditions,~\eqref{subeq:sysDynamic} and~\eqref{subeq:outputMap} express the discretized dynamic model for the~\ac{GTMR} and the output signals of the system, respectively, and actuator limits ($\underline{\gamma}, \bar{\gamma}, \underline{\dot{\gamma}}, \bar{\dot{\gamma}}$) are embedded in~\eqref{subeq:uBound} and~\eqref{subeq:dotuBound}. The constraints~\eqref{sueq:misAligConstr} ensure that~\ac{UAV}-1 will be aligned to~\ac{UAV}-2 and the~\ac{BS} while moving. The variable $\mathcal{T}$ refers to communication parameters that need to be taken into consideration while solving the problem. Finally, the vectors $\bar{\mathbf{u}}_k$, $\bar{\mathbf{x}}_k$, $\mathbf{y}_{\mathrm{d},k}$, and $\mathbf{y}_k$ denote the $k$-th element of vectors $\bar{\mathbf{u}}$, $\bar{\mathbf{x}}$, $\mathbf{y}_{\mathrm{d}}$, and $\mathbf{y}$, respectively. The feasibility and effectiveness of the control strategy has been demonstrated via closed-loop simulations achieved in MATLAB, as depicted in Fig.~\ref{fig:comparisonBitTransmitted}. The proposed control strategy~\eqref{eq:NMPC_formulation} follows the problem in~\eqref{TPA:eq:1}, embedding both communication and robot dynamics constraints in the optimal problem formulation.

\section{Research Opportunities}
\label{sec:researchOpportunities}

There is increasing interest in the integration of communication systems with~\acp{UGV} and~\acp{UAV}. A multidisciplinary approach is necessary to fully realize the potential of this field, as there is a strong connection between the robotics and communication aspects. Some current research topics in this area are the following.

\textbf{Communication-aware trajectory for~\ac{PV} equipped robots}: there has been recent interest in adding solar~\ac{PV} panels to~\acp{UAV} in order to extend their flight time and increase mission endurance. However, the addition of~\ac{PV} panels also introduces a new element to the~\ac{CaTP} problem, as the solar power production of the~\ac{UAV}'s~\ac{PV} panels and the quality of its communication are both functions of the~\ac{UAV}'s state. Thus, in addition to considering the motion and communication aspects, the energy generation aspect must also be taken into account~\cite{Sun2019IEEETC}.

\textbf{\acp{UAV} and physical security}: the use of~\acp{UAV} to address cybersecurity issues in cellular networks through physical layer techniques is an emerging and interesting application. One example is the joint optimization of a~\ac{UAV}'s trajectory and communication to improve network security against eavesdroppers. This can be achieved through the use of one or more~\acp{UAV} to (i) jam the eavesdropper to degrade its~\ac{SNR} or (ii) act as relays between the base station and the targeted node, allowing the latter to reduce its transmission power and decreasing the strength of signals captured by eavesdroppers~\cite{Liu2019IEEETWC}.

\textbf{\ac{FSO} for~\acp{UAV}}:
they are currently being integrated into ~\ac{UAV} communications \cite{DabiriIEEECL2020, LeeIEEETWC2020, DabiriIEEEJSAC2018}, uses a laser to transmit data and a~\ac{PD} to receive it. One advantage over~\ac{RF} communications is its immunity against~\ac{RF} interference and are therefore also resistant to electromagnetic jamming. Additionally, physically blocking the~\ac{LoS} between two~\acp{UAV} using~\ac{FSO} communications is extremely difficult, and if it does happen it would be easily noticed by both~\acp{UAV}. But,~\ac{FSO} communications are strongly affected by misalignment between the laser and~\ac{PD}, which is exacerbated by the vibrations of~\acp{UAV} and environmental perturbations, such as wind. To fully realize the benefits of~\ac{UAV}~\ac{FSO} communications, it is necessary to consider the dynamic behavior of~\acp{UAV}. The mathematical structure described in Sec.~\ref{sec:communicationAwareTrajectory} can also be used for~\ac{CaTP} for~\ac{FSO} communications in~\acp{UAV} by simply replacing the~\ac{RF} communication channel models with the corresponding optical communication channel models.

Finally, an extended version of this paper is available in~\cite{licea2022robotics}.




\bibliographystyle{IEEEtran}
\bibliography{bib_short.bib}

\begin{thebibliography}{10}
\providecommand{\url}[1]{#1}
\csname url@samestyle\endcsname
\providecommand{\newblock}{\relax}
\providecommand{\bibinfo}[2]{#2}
\providecommand{\BIBentrySTDinterwordspacing}{\spaceskip=0pt\relax}
\providecommand{\BIBentryALTinterwordstretchfactor}{4}
\providecommand{\BIBentryALTinterwordspacing}{\spaceskip=\fontdimen2\font plus
\BIBentryALTinterwordstretchfactor\fontdimen3\font minus
  \fontdimen4\font\relax}
\providecommand{\BIBforeignlanguage}[2]{{%
\expandafter\ifx\csname l@#1\endcsname\relax
\typeout{** WARNING: IEEEtran.bst: No hyphenation pattern has been}%
\typeout{** loaded for the language `#1'. Using the pattern for}%
\typeout{** the default language instead.}%
\else
\language=\csname l@#1\endcsname
\fi
#2}}
\providecommand{\BIBdecl}{\relax}
\BIBdecl

\bibitem{Gasparri2017TRO}
A.~Gasparri \emph{et~al.}, ``{Bounded Control Law for Global Connectivity
  Maintenance in Cooperative Multirobot Systems},'' \emph{IEEE Transactions on
  Robotics}, vol.~33, no.~3, pp. 700--717, 2017.

\bibitem{Licea2020TRO}
D.~Licea~Bonilla \emph{et~al.}, ``{Communication-Aware Energy Efficient
  Trajectory Planning With Limited Channel Knowledge},'' \emph{IEEE
  Transactions on Robotics}, vol.~36, no.~2, pp. 431--442, 2020.

\bibitem{Zeng2017TWC}
Y.~Zeng \emph{et~al.}, ``{Energy-Efficient UAV Communication With Trajectory
  Optimization},'' \emph{IEEE Transactions on Wireless Communications},
  vol.~16, no.~6, pp. 3747--3760, 2017.

\bibitem{Wu2018TWC}
Q.~Wu \emph{et~al.}, ``{Joint Trajectory and Communication Design for Multi-UAV
  Enabled Wireless Networks},'' \emph{IEEE Transactions on Wireless
  Communications}, vol.~17, no.~3, pp. 2109--2121, 2018.

\bibitem{Zeng2019ProceedingsIEEE}
Y.~Zeng \emph{et~al.}, ``Accessing from the sky: A tutorial on uav
  communications for 5g and beyond,'' \emph{Proceedings of the IEEE}, vol. 107,
  no.~12, pp. 2327--2375, 2019.

\bibitem{Ahmad2022swarms}
A.~{Ahmad} \emph{et~al.}, ``{PACNav: A Collective Navigation Approach for UAV
  Swarms Deprived of Communication and External Localization},''
  \emph{Bioinspiration \& Biomimetics}, vol.~17, no.~6, p. 066019, 2022.

\bibitem{Jung2010CM}
J.~H. Jung \emph{et~al.}, ``{Multi-robot path finding with wireless multihop
  communications},'' \emph{IEEE Communications Magazine}, vol.~48, no.~7, pp.
  126--132, 2010.

\bibitem{LindheICRA2010}
M.~Lindhé \emph{et~al.}, ``{Adaptive exploitation of multipath fading for
  mobile sensors},'' in \emph{2010 IEEE International Conference on Robotics
  and Automation}, 2010, pp. 1934--1939.

\bibitem{CalvoFullana2021IEEECM}
M.~Calvo-Fullana \emph{et~al.}, ``{Communications and Robotics Simulation in
  UAVs: A Case Study on Aerial Synthetic Aperture Antennas},'' \emph{IEEE
  Communications Magazine}, vol.~59, no.~1, pp. 22--27, 2021.

\bibitem{CalvoFullana2021RAL}
------, ``{ROS-NetSim: A Framework for the Integration of Robotic and Network
  Simulators},'' \emph{IEEE Robotics and Automation Letters}, vol.~6, no.~2,
  pp. 1120--1127, 2021.

\bibitem{Guo2018TRO}
M.~Guo \emph{et~al.}, ``{Multirobot Data Gathering Under Buffer Constraints and
  Intermittent Communication},'' \emph{IEEE Transactions on Robotics}, vol.~34,
  no.~4, pp. 1082--1097, 2018.

\bibitem{Siciliano2016Handbook}
B.~Siciliano \emph{et~al.}, \emph{{Springer Handbook of Robotics}},
  2nd~ed.\hskip 1em plus 0.5em minus 0.4em\relax Springer, 2016, ch.~26, pp.
  623--670.

\bibitem{HammoutiUnet2018}
H.~{El Hammouti} \emph{et~al.}, ``{Air-to-Ground Channel Modeling for UAV
  Communications Using 3D Building Footprints},'' in \emph{Ubiquitous
  Networking}.\hskip 1em plus 0.5em minus 0.4em\relax Springer International
  Publishing, 2018, pp. 372--383.

\bibitem{uav15}
A.~A. {Khuwaja} \emph{et~al.}, ``{A Survey of Channel Modeling for UAV
  Communications},'' \emph{IEEE Communications Surveys Tutorials}, vol.~20,
  no.~4, pp. 2804--2821, 2018.

\bibitem{UAVChannelsurvey2}
W.~{Khawaja} \emph{et~al.}, ``{A Survey of Air-to-Ground Propagation Channel
  Modeling for Unmanned Aerial Vehicles},'' \emph{IEEE Communications Surveys
  Tutorials}, vol.~21, no.~3, pp. 2361--2391, 2019.

\bibitem{boyd_vandenberghe_2004}
S.~Boyd \emph{et~al.}, \emph{{Convex Optimization}}.\hskip 1em plus 0.5em minus
  0.4em\relax Cambridge University Press, 2004.

\bibitem{Silano2021RAL}
G.~Silano \emph{et~al.}, ``{Power Line Inspection Tasks With Multi-Aerial Robot
  Systems Via Signal Temporal Logic Specifications},'' \emph{IEEE Robotics and
  Automation Letters}, vol.~6, no.~2, pp. 4169--4176, 2021.

\bibitem{HertICUAS2022}
D.~Hert \emph{et~al.}, ``{MRS Modular UAV Hardware Platforms for Supporting
  Research in Real-World Outdoor and Indoor Environments},'' in \emph{2022
  International Conference on Unmanned Aircraft Systems}, 2022, pp. 1264--1273.

\bibitem{SilanoICUAS2021}
G.~Silano \emph{et~al.}, ``{A Multi-Layer Software Architecture for Aerial
  Cognitive Multi-Robot Systems in Power Line Inspection Tasks},'' in
  \emph{2021 International Conference on Unmanned Aircraft Systems}, 2021, pp.
  1624--1629.

\bibitem{CalvoICUAS2022-BT}
A.~Calvo \emph{et~al.}, ``{Mission Planning and Execution in Heterogeneous
  Teams of Aerial Robots supporting Power Line Inspection Operations},'' in
  \emph{2022 International Conference on Unmanned Aircraft Systems}, 2022, pp.
  1644--1649.

\bibitem{Licea2021EUSIPCO}
D.~Bonilla~Licea \emph{et~al.}, ``{Optimum Trajectory Planning for Multi-Rotor
  UAV Relays with Tilt and Antenna Orientation Variations},'' in \emph{2021
  29th European Signal Processing Conference}, 2021, pp. 1586--1590.

\bibitem{Andriy2022ICUAS}
A.~{Dmytruk} \emph{et~al.}, ``{A Perception-Aware NMPC for Vision-Based Target
  Tracking and Collision Avoidance with a Multi-Rotor UAV},'' in \emph{2022
  International Conference on Unmanned Aircraft Systems}, 2022, pp. 1668--1673.

\bibitem{CataffoSMC2022}
V.~Cataffo \emph{et~al.}, ``{A Nonlinear Model Predictive Control Strategy for
  Autonomous Racing of Scale Vehicles},'' in \emph{2022 IEEE International
  Conference on Systems, Man, and Cybernetics}, 2022, pp. 100--105.

\bibitem{Michieletto2018TRO}
G.~Michieletto \emph{et~al.}, ``{Fundamental Actuation Properties of
  Multirotors: Force–Moment Decoupling and Fail–Safe Robustness},''
  \emph{IEEE Transactions on Robotics}, vol.~34, no.~3, pp. 702--715, 2018.

\bibitem{Sun2019IEEETC}
Y.~Sun \emph{et~al.}, ``{Optimal 3D-Trajectory Design and Resource Allocation
  for Solar-Powered UAV Communication Systems},'' \emph{IEEE Transactions on
  Communications}, vol.~67, no.~6, pp. 4281--4298, 2019.

\bibitem{Liu2019IEEETWC}
C.~Liu \emph{et~al.}, ``{Safeguarding UAV Communications Against Full-Duplex
  Active Eavesdropper},'' \emph{IEEE Transactions on Wireless Communications},
  vol.~18, no.~6, pp. 2919--2931, 2019.

\bibitem{DabiriIEEECL2020}
M.~T. Dabiri \emph{et~al.}, ``{Optimal Placement of UAV-Assisted Free-Space
  Optical Communication Systems With DF Relaying},'' \emph{IEEE Communications
  Letters}, vol.~24, no.~1, pp. 155--158, 2020.

\bibitem{LeeIEEETWC2020}
J.-H. Lee \emph{et~al.}, ``{A UAV-Mounted Free Space Optical Communication:
  Trajectory Optimization for Flight Time},'' \emph{IEEE Transactions on
  Wireless Communications}, vol.~19, no.~3, pp. 1610--1621, 2020.

\bibitem{DabiriIEEEJSAC2018}
M.~T. Dabiri \emph{et~al.}, ``{Channel Modeling and Parameter Optimization for
  Hovering UAV-Based Free-Space Optical Links},'' \emph{IEEE Journal on
  Selected Areas in Communications}, vol.~36, no.~9, pp. 2104--2113, 2018.

\bibitem{licea2022robotics}
D.~{Bonilla Licea} \emph{et~al.}, ``{When Robotics Meets Wireless
  Communications: An Introductory Tutorial},'' \emph{ArXiv, 2209.02021}, 2022.

\end{thebibliography}

\end{document}